\pdfoutput=1

\documentclass[11pt]{article}

\usepackage[]{EMNLP2023}

\usepackage{CJK}
\usepackage{times}
\usepackage{latexsym}

\usepackage[T1]{fontenc}

\usepackage[utf8]{inputenc}

\usepackage{microtype}

\usepackage{inconsolata}

\usepackage{tabularx}
\usepackage{booktabs,tabularx}
\usepackage{todonotes}
\usepackage[capitalise,noabbrev]{cleveref}
\usepackage{listings}
\usepackage{fancyvrb} 

\definecolor{verbgray}{gray}{0.9}

\usepackage{colortbl}
\definecolor{lightgray}{rgb}{0.7,0.7,0.7}
\def\grayrow{\rowcolor{lightgray}}

\def\parcite#1{\citep{#1}} 
\def\perscite#1{\citet{#1}} 
\def\inparcite#1{\citealp{#1}} 

\title{GEMBA-MQM: Detecting Translation Quality Error Spans with GPT-4}

\author{Tom Kocmi \and Christian Federmann \\
  Microsoft,
  One Microsoft Way,
  Redmond, WA-98052, USA \\
  \texttt{\{tomkocmi,chrife\}@microsoft.com}} 

\begin{document}
\maketitle
\begin{abstract}

This paper introduces GEMBA-MQM, a GPT-based evaluation metric designed to detect translation quality errors, specifically for the quality estimation setting without the need for human reference translations. Based on the power of large language models (LLM), GEMBA-MQM employs a fixed three-shot prompting technique, querying the GPT-4 model to mark error quality spans. Compared to previous works, our method has language-agnostic prompts, thus avoiding the need for manual prompt preparation for new languages.

While preliminary results indicate that GEMBA-MQM achieves state-of-the-art accuracy for system ranking, we advise caution when using it in academic works to demonstrate improvements over other methods due to its dependence on the proprietary, black-box GPT model.
\end{abstract}

\begin{table}[!htb]
\small
\centering
\begin{tabular}{lll}
Metric                         & Acc. & Meta \\
\midrule
GEMBA-MQM           & 96.5\% (1)   & 0.802 (3)  \\
\grayrow{}  XCOMET-Ensemble                & 95.2\% (1)   & 0.825 (1)  \\
\grayrow{}  docWMT22CometDA                & 93.7\% (2)   & 0.768 (9)  \\
docWMT22CometKiwiDA & 93.7\% (2)   & 0.767 (9)  \\
XCOMET-QE-Ensemble  & 93.5\% (2)   & 0.808 (2)  \\
\grayrow{}  COMET                          & 93.5\% (2)   & 0.779 (6)  \\
\grayrow{}  MetricX-23                     & 93.4\% (3)   & 0.808 (2)  \\
CometKiwi           & 93.2\% (3)   & 0.782 (5)  \\
\grayrow{}  Calibri-COMET22                & 93.1\% (3)   & 0.767 (10) \\
\grayrow{}  BLEURT-20                      & 93.0\% (4)     & 0.776 (7)  \\
\grayrow{}  MaTESe                         & 92.8\% (4)   & 0.782 (5)  \\
\grayrow{}  mre-score-labse-regular        & 92.7\% (4)   & 0.743 (13) \\
mbr-bleurtxv1p-qe   & 92.5\% (4)   & 0.788 (4)  \\
KG-BERTScore        & 92.5\% (5)   & 0.774 (7)  \\
MetricX-23-QE       & 92.0\% (5)     & 0.800 (3)    \\
\grayrow{}  BERTscore                      & 90.2\% (7)   & 0.742 (13) \\
MS-COMET-QE-22      & 90.1\% (8)   & 0.744 (12) \\
\grayrow{}  embed\_llama                   & 87.3\% (10)  & 0.701 (16) \\
\grayrow{}  f200spBLEU                     & 86.8\% (11)  & 0.704 (15) \\
\grayrow{}  BLEU                           & 85.9\% (12)  & 0.696 (16) \\
\grayrow{}  chrF                           & 85.2\% (12)  & 0.694 (17)
\end{tabular}
\caption{Preliminary results of the WMT 2023 Metric Shared task. The first column shows the system-level accuracy, and the second column is the Metrics 2023 meta evaluation. Metrics with gray background need human references. The table does not contain the worst-performing, non-standard metrics due to space reasons.}
\label{tab:wmt23results}
\vspace{-5mm}
\end{table}

\begin{figure*}[htb]
{\footnotesize
    \begin{Verbatim}[commandchars=+\[\]]
    (System) You are an annotator for the quality of machine translation. Your task is to identify 
    errors and assess the quality of the translation.
    
    (user) +textbf[{source_language}] source:\n
    ```+textbf[{source_segment}]```\n
    +textbf[{target_language}] translation:\n
    ```+textbf[{target_segment}]```\n
    \n
    Based on the source segment and machine translation surrounded with triple backticks, identify 
    error types in the translation and classify them. The categories of errors are: accuracy 
    (addition, mistranslation, omission, untranslated text), fluency (character encoding, grammar, 
    inconsistency, punctuation, register, spelling), 
    +highlight[locale convention (currency, date, name, telephone, or time format)] 
    style (awkward), terminology (inappropriate  for context, inconsistent use), non-translation, 
    other, or no-error.\n
    Each error is classified as one of three categories: critical, major, and minor. 
    Critical errors inhibit comprehension of the text. Major errors disrupt the flow, but what 
    the text is trying to say is still understandable. Minor errors are technically errors, 
    but do not disrupt the flow or hinder comprehension.

    (assistant) +textbf[{observed error classes}]
    \end{Verbatim}
}
\caption{The general prompt for GEMBA-MQM omits the gray part which performed subpar on internal data (we include it in GEMBA-locale-MQM). The ``(user)'' and ``(assistant)'' section is repeated for each few-shot example.}
\label{fig:prompt}
\end{figure*}

\section{Introduction}

GEMBA-MQM builds on the recent finding that large language models (LLMs) can be prompted to assess the quality of machine translation \parcite{kocmi2023large}.

The earlier work \perscite{kocmi2023large} (GEMBA-DA) adopted a straightforward methodology of assessing single score values for each segment without specifying the scale in detail. Employing a zero-shot approach, their technique showed an unparalleled accuracy in assessment, surpassing all other non-LLM metrics on the WMT22 metrics test set \parcite{freitag-etal-2022-results}.

Next, \perscite{Lu2023EAPrompt} (EAPrompt) investigated prompting LLMs to assess individual error classes from a multidimensional quality metrics (MQM) framework \parcite{freitag-etal-2021-experts}, where each error can be classified into various error classes (such as accuracy, fluency, style, terminology, etc.), subclasses (accuracy > mistranslation), and is marked with its severity (critical, major, minor). Segment scores are computed by aggregating errors, each weighted by its respective severity coefficient (25, 5, 1).
While their approach employed a few-shot prompting with a chain-of-thought strategy \parcite{wei2022chain}, our GEMBA-MQM approach differs in two aspects: 1) We streamline the process using only single-step prompting, and 2) our prompts are universally applicable across languages, avoiding the need for manual prompt preparation for each language pair.

Another notable effort by \perscite{fernandes2023devil} paralleled the EAPrompt approach, also marking MQM error spans. In contrast, their approach used a PaLM-2 model, pooling MQM annotations to sample a few shot examples for the prompt. Their fine-tuning experiments did not improve system-level performance for the top-tier models.

\section{Description}

Our technique adopts few-shot learning with the GPT-4 model \parcite{openai2023gpt4}, prompting the model to mark quality error spans using the MQM framework. The underlying prompt template is modeled on guidelines for human annotators and shown in \cref{fig:prompt}.

In contrast to other methods, we use three pre-determined examples (see \cref{app:examples}), allowing the method to be used with any language pair, avoiding the need to create language pair specific MQM few-shot examples. This was the original limitation that prevented \perscite{fernandes2023devil} from evaluating AutoMQM beyond two language pairs. Our decision was not driven by a desire to enhance performance — since domain and language-specific prompts typically boost it \parcite{moslem-etal-2023-adaptive} — but rather to ensure our method can be evaluated across any language pairs.

\section{Experiments}

To measure the performance of the GEMBA-MQM metric, we follow the methodology and use test data provided by the WMT22 Metrics shared task \parcite{freitag-etal-2022-results} which hosts an annual evaluation of automatic metrics, benchmarking them against human gold labels. 

We compare our method against the best-performing reference-based metrics of WMT22: MetrixX\_XXL (non-public metric), COMET-22 \parcite{rei-etal-2022-comet}, UNITE \parcite{wan-etal-2022-unite}, BLEURT-20 \parcite{pu-etal-2021-learning}, and COMET-20 \parcite{rei-etal-2020-comet}. In addition, we also compare against ``classic'' string-based metrics BLEU \parcite{papineni-etal-2002-bleu} and ChrF \parcite{popovic-2015-chrf}.
Lastly, we compare against reference-less metrics of WMT22: CometKIWI \parcite{rei-etal-2022-comet}, Unite-src \parcite{wan-etal-2022-alibaba}, Comet-QE \parcite{rei-etal-2021-references}, MS-COMET-QE-22 \parcite{kocmi-etal-2022-ms}.

We contrast our work with other LLM-based evaluation methods such as GEMBA-DA \parcite{kocmi-federmann-2023-large} and EAPrompt \parcite{Lu2023EAPrompt}, conducting experiments using two GPT models: GPT-3.5-Turbo and the more powerful GPT-4 \parcite{openai2023gpt4}.

\subsection{Test set}

The main evaluation of our work has been done on the MQM22 \parcite{freitag-etal-2022-results} and internal Microsoft data. Furthermore, a few days before the camera-ready deadline, organizers of Metrics 2023 \parcite{metrics2023} released results on the blind test set, showing performance on unseen data.

The MQM22 test set contains human judgments for three translation directions: English into German, English into Russian, and Chinese into English. The test set contains a total of 54 machine translation system outputs or human translations. It contains a total of 106k segments. Translation systems are mainly from participants of the WMT22 General MT shared task \parcite{kocmi-etal-2022-findings}. The source segments and human reference translations for each language pair contain around 2,000 sentences from four different text domains: news, social, conversational, and e-commerce. The gold standard for scoring translation quality is based on human MQM ratings, annotated by professionals who mark individual errors in each translation, as described in \perscite{freitag-etal-2021-experts}.

The MQM23 test set is the blind set for this year's WMT Metrics shared task prepared in the same way as MQM22, but with unseen data for all participants, making it the most reliable evaluation as neither participants nor LLM could overfit to those data. The main difference from last year's iteration is the replacement of English into Russian with Hebrew into English. Also, some domains have been updated; see \perscite{generalmt2023}.

Additionally, we evaluated GEMBA-MQM on a large internal test set, an extended version of the data set described by \perscite{kocmi-etal-2021-ship}. This test set contains human scores collected with source-based Direct Assessment (DA, \inparcite{graham-etal-2013-continuous}) and its variant DA+SQM \parcite{kocmi-etal-2022-findings}. This test set contains 15 high-resource languages paired with English. Specifically, these are: Arabic, Czech, Dutch, French, German, Hindi, Italian, Japanese, Korean, Polish, Portuguese, Russian, Simplified Chinese, Spanish, and Turkish.

\subsection{Evaluation methods}

The main use case of automatic metrics is system ranking, either when comparing a baseline to a new model, when claiming state-of-the-art results, when comparing different model architectures in ablation studies, or when deciding if to deploy a new model to production. Therefore, we focus on a method that specifically measures this target: system-level pairwise accuracy \parcite{kocmi-etal-2021-ship}.

The pairwise accuracy is defined as the number of system pairs ranked correctly by the metric with respect to the human ranking divided by the total number of system pair comparisons.

Formally:

\begin{center}
\footnotesize
\[\mbox{Accuracy} = \frac{|\mbox{sign}(\mbox{metric} \Delta)==\mbox{sign}(\mbox{human} \Delta)|}{|\mbox{all system pairs}|}\]
\end{center}

We reproduced all scores reported in the WMT22 Metrics shared task findings paper using the official WMT22 script.\footnote{\scriptsize\url{https://github.com/google-research/mt-metrics-eval}} Reported scores match Table 11 of the WMT22 metrics findings paper \parcite{freitag-etal-2022-results}.

Furthermore, organizers of Metrics shared task 2023 defined a new meta-evaluation metric based on four different scenarios, each contributing to the final score with a weight of 0.25:

\begin{itemize}
    \setlength\itemsep{-0.5em}
    \item[--] system-level pairwise accuracy;
    \item[--] system-level Pearson correlation;
    \item[--] segment-level Accuracy-t \parcite{deutsch2023ties}; and
    \item[--] segment-level Pearson correlation.
\end{itemize}

The motivation is to measure metrics in the most general usage scenarios (for example, for segment-level filtering) and not just for system ranking. However, we question the decision behind the use of Pearson correlation, especially on the system level. As \perscite{mathur-etal-2020-tangled} showed, Pearson used for metric evaluation is sensitive when applied to small sample sizes (in MQM23, the sample size is as little as 12 systems); it is heavily affected by outliers \parcite{osborne2004power, ma-etal-2019-results}, which need to be removed before running the evaluation; and it measures linear correlation with the gold MQM data, which are not necessarily linear to start with (especially the discrete segment-level scores, with error weights of 0.1, 1, 5, 25).

Although it is desirable to have an automatic metric that correlates highly with human annotation behaviour and which is useful for segment-level evaluation, more research is needed regarding the proper way of testing these properties.

\begin{table}
\small
\centering
\begin{tabular}{ll}
Metric                                & Acc.     \\
\midrule
\grayrow{}  EAPrompt-Turbo      & 90.9\%  \\
\grayrow{}  GEMBA-DA-GPT4       & 89.8\%  \\
 GEMBA-locale-MQM-Turbo   & 89.8\% \\
 EAPrompt-Turbo                 & 89.4\%  \\
 GEMBA-MQM-GPT4                 & 89.4\%  \\
 GEMBA-DA-GPT4                  & 87.6\%  \\
 GEMBA-DA-Turbo                 & 86.9\%  \\
 GEMBA-MQM-Turbo                & 86.5\%  \\
\grayrow{}  GEMBA-DA-Turbo      & 86.5\%  \\
\grayrow{}  MetricX\_XXL        & 85.0\%  \\
\grayrow{}  BLEURT-20           & 84.7\%  \\
\grayrow{} COMET-22            & 83.9\%  \\
\grayrow{}  COMET-20            & 83.6\%  \\
\grayrow{}  UniTE               & 82.8\%  \\
COMETKiwi                      & 78.8\%  \\
 COMET-QE                       & 78.1\%  \\
\grayrow{} BERTScore           & 77.4\%  \\
 UniTE-src                      & 75.9\%  \\
 MS-COMET-QE-22                 & 75.5\%  \\
\grayrow{} chrF                & 73.4\%  \\
\grayrow{} BLEU                & 70.8\% 
\end{tabular}
\caption{The system-level pairwise accuracy results for the WMT 22 metrics task test set. Gray metrics need reference translations which are not the focus of the current evaluation.}
\label{tab:results_mqm}
\end{table}

\section{Results}

In this section, we discuss the results observed on three different test sets: 1) MQM test data from WMT, 2) internal test data from Microsoft, and 3) a subset of the internal test data to measure the impact of the MQM locale convention.

\subsection{Results on MQM Test Data from WMT}

The results of the blind set MQM23 in \cref{tab:wmt23results} show that GEMBA-MQM outperforms all other techniques on the three languages evaluated in the system ranking scenario. Furthermore, when evaluated in the meta-evaluation scenario it achieves the third cluster rank.

In addition to the official results, we also test on MQM22 test data and show results in \cref{tab:results_mqm}. The main conclusion is that all GEMBA-MQM variants outperform traditional metrics (such as COMET or Metric XXL). When focusing on the quality estimation task, we can see that the GEMBA-locale-MQM-Turbo method slightly outperforms EAPrompt, which is the closest similar technique.

However, we can see that our final technique GEMBA-MQM is performing significantly worse than the GEMBA-locale-MQM metric, while the only difference is the removal of the locale convention error class. We believe this to be caused by the test set. We discuss our decision to remove the locale convention error class in \cref{sec:removal_locale}.

\subsection{Results on Internal Test Data}

\cref{tab:internal_results} shows that GEMBA-MQM-Turbo outperforms almost all other metrics, losing only to COMETKIWI-22. This shows some limitations of GPT-based evaluation on blind test sets. Due to access limitations, we do not have results for GPT-4, which we assume should outperform the GPT-3.5 Turbo model. We leave this experiment for future work.

\begin{table}
\small
\centering
\begin{tabular}{lrr}
                       & 15 langs & Cs + De \\
\# of system pairs (N) & 4,468    & 734     \\
\midrule
COMETKiwi              & 79.9     & 81.3    \\
GEMBA-locale-MQM-Turbo & 78.6     & 81.3    \\
GEMBA-MQM-Turbo        & 78.4     & 83.0    \\
COMET-QE               & 77.8     & 79.8    \\
\grayrow{}  COMET-22   & 76.5     & 79.2    \\
\grayrow{}  COMET-20   & 76.3     & 79.6    \\
\grayrow{} BLEURT-20   & 75.8     & 79.7    \\
\grayrow{} chrF        & 68.1     & 70.6    \\
\grayrow{} BLEU        & 66.8     & 68.9   
\end{tabular}
\caption{System-level pairwise accuracy results for our internal test set. The first column is for all 15 languages, and the second is Czech and German only. All languages are paired with English.}
\label{tab:internal_results}
\end{table}

\subsection{Removal of Locale Convention}
\label{sec:removal_locale}

When investigating the performance of GEMBA-locale-MQM on a subset of internal data (Czech and German), we observed a critical error in this prompt regarding the "locale convention" error class. GPT assigned this class for errors not related to translations. It flagged Czech sentences as a locale convention error when the currency Euro was mentioned, even when the translation was fine, see example in \cref{tab:locale_problem}. We assume that it was using this error class to mark parts not standard for a given language but more investigation would be needed to draw any deeper conclusions.

\begin{table}
\small
\begin{tabular}{ll}
Source          & Vstupné do památky činí 16,50 Eur.\\
Hypothesis      & Admission to the monument is 16.50 Euros.\\
GPT annot. & locale convention/currency: "euros" 
\end{tabular}
\caption{An example of a wrong error class ``locale convention'' as marked by GEMBA-locale-MQM. The translation is correct, however, we assume that the GPT model might not have liked the use of Euros in a Czech text because Euros are not used in the Czech Republic.}
\label{tab:locale_problem}
\end{table}

\begin{table}[t]
\small
\begin{tabular}{lrr}
Error class       & GEMBA-locale-MQM & GEMBA-MQM \\
\midrule
accuracy          & 960.838 (39\%) & 1.072.515 (51\%) \\
locale con. & 808.702 (32\%) & (0\%)          \\
fluency           & 674.228 (27\%) & 699.037 (33\%)  \\
style             & 23.943 (1\%)   & 41.188 (2\%)    \\
terminology       & 17.379 (1\%)   & 290.490 (14\%)  \\
Other errors    & 4.126 (0\%)    & 10615 (1\%)    \\
\midrule
Total             & 2.489.216       & 2.113.845       
\end{tabular}
\caption{Distribution of errors for both types of prompts over all segments of the internal test set for the Turbo model.}
\label{tab:error_distribution}
\end{table}

The evaluation on internal test data in \cref{tab:locale_problem} showed gains of 1.7\% accuracy. However, when evaluating over 15 languages, we observed a small degradation of 0.2\%. For MQM22 in \cref{tab:results_mqm}, the degradation is even bigger.

When we look at the distribution of the error classes over the fifteen highest resource languages in \cref{tab:error_distribution}, we observe that 32\% of all errors for GEMBA-locale-MQM are marked as a locale convention suggesting a misuse of GPT for this error class. Therefore, instead of explaining this class in the prompt, we removed it. This resulted in about half of the original locale errors being reassigned to other error classes, while the other half was not marked.

In conclusion, we decided to remove this class as it is not aligned with what we expected to measure and how GPT appears to be using the classes. Thus, we force GPT to classify those errors using other error categories. Given the different behaviour for internal and external test data, this deserves more investigation in future work.



\section{Caution with ``Black Box'' LLMs}

Although GEMBA-MQM is the state-of-the-art technique for system ranking, we would like to discuss in this section the inherent limitations of using ``black box'' LLMs (such as GPT-4) when conducting academic research.

Firstly, we would like to point out that GPT-4 is a proprietary model, which leads to several problems. One of them is that we do not know which training data it was trained on, therefore any published test data should be considered as part of their training data (and is, therefore, possibly tainted).
Secondly, we cannot guarantee that the model will be available in the future, or that it won't be updated in the future, meaning any results from such a model are relevant only for the specific sampling time. As \perscite{chen2023chatgpt} showed, the model's performance fluctuated and decreased over the span of 2023.

As this impacts all proprietary LLMs, we advocate for increased research using publicly available models, like LLama 2 \parcite{touvron2023llama}. This approach ensures future findings can be compared both to ``black box'' LLMs while also allowing comparison to ``open'' models.\footnote{Although LLama 2 is not fully open, its binary files have been released. Thus, when used it as a scorer, we are using the exact same model.}

\section{Conclusion}

In this paper, we have introduced and evaluated the GEMBA-MQM metric, a GPT-based metric for translation quality error marking. This technique takes advantage of the GPT-4 model with a fixed three-shot prompting strategy. 
Preliminary results show that GEMBA-MQM achieves a new state of the art when used as a metric for system ranking, outperforming established metrics such as COMET and BLEURT-20. 

We would like to acknowledge the inherent limitations tied to using a proprietary model like GPT. Our recommendation to the academic community is to be cautious with employing GEMBA-MQM on top of GPT models. For future research, we want to explore how our approach performs with other, more open LLMs such as LLama 2 \parcite{touvron2023llama}. Confirming superior behaviour on publicly distributed models (at least their binaries) could open the path for broader usage of the technique in the academic environment.

\section*{Limitations}

While our findings and techniques with GEMBA-MQM bring promising advancements in translation quality error marking, it is essential to highlight the limitations encountered in this study.

\begin{itemize}

\item[--] Reliance on Proprietary GPT Models: GEMBA-MQM depends on the GPT-4 model, which remains proprietary in nature. We do not know what data the model was trained on or if the same model is still deployed and therefore the results are comparable. As \perscite{chen2023chatgpt} showed, the model's performance fluctuated throughout 2023;

\item[--] High-Resource Languages Only: As WMT evaluations primarily focus on high-resource languages, we cannot conclude if the method will perform well on low-resource languages.

\end{itemize}

\section*{Acknowledgements}

We are grateful to our anonymous reviewers for their insightful comments and patience that have helped improve the paper. We would like to thank our colleagues on the Microsoft Translator research team for their valuable feedback.

\bibliography{anthology,custom}
\bibliographystyle{acl_natbib}

\appendix
\clearpage
\onecolumn

\section{Three examples Used for Few-shot Prompting}
\label{app:examples}

\begin{figure*}[htb]
{\footnotesize
    \begin{Verbatim}
English source: I do apologise about this, we must gain permission from the account holder to discuss
an order with another person, I apologise if this was done previously, however, I would not be able 
to discuss this with yourself without the account holders permission.
German translation: Ich entschuldige mich dafür, wir müssen die Erlaubnis einholen, um eine Bestellung 
mit einer anderen Person zu besprechen. Ich entschuldige mich, falls dies zuvor geschehen wäre, aber 
ohne die Erlaubnis des Kontoinhabers wäre ich nicht in der Lage, dies mit dir involvement.
MQM annotations: 
Critical: 
no-error
Major:
accuracy/mistranslation - "involvement"
accuracy/omission - "the account holder"
Minor:
fluency/grammar - "wäre"
fluency/register - "dir"
    \end{Verbatim}
    
    \begin{Verbatim}
English source: Talks have resumed in Vienna to try to revive the nuclear pact, with both sides 
trying to gauge the prospects of success after the latest exchanges in the stop-start negotiations.
Czech transation: Ve Vídni se ve Vídni obnovily rozhovory o oživení jaderného paktu, přičemže obě 
partaje se snaží posoudit vyhlídky na úspěch po posledních výměnách v jednáních.
MQM annotations: 
Critical:
no-error
Major:
accuracy/addition - "ve Vídni"
accuracy/omission - "the stop-start"
Minor:
terminology/inappropriate for context - "partaje"
    \end{Verbatim}
    
\begin{CJK}{UTF8}{gbsn}
    \begin{Verbatim}
Chinese source: 大众点评乌鲁木齐家居商场频道为您提供高铁居然之家地址，电话，营业时间等最新商户信息，
找装修公司，就上大众点评
English translation: Urumqi Home Furnishing Store Channel provides you with the latest business 
information such as the address, telephone number, business hours, etc., of high-speed rail, and 
find a decoration company, and go to the reviews.
MQM annotations: 
Critical:
accuracy/addition - "of high-speed rail"
Major:
accuracy/mistranslation - "go to the reviews"
Minor:
style/awkward - "etc.,"
    \end{Verbatim}
\end{CJK}
}
\caption{Three examples used for all languages.}
\end{figure*}

\end{document}